\documentclass[conference]{IEEEtran}
\IEEEoverridecommandlockouts
\usepackage{url}
\usepackage{amsmath,amssymb,amsfonts}
\usepackage[backend=biber,style=ieee, maxnames=5]{biblatex}
\usepackage{algorithmic}
\usepackage{graphicx}
\usepackage{textcomp}
\usepackage{listings}
\usepackage{pdflscape}
\usepackage{multirow}
\usepackage{booktabs}

\usepackage{multirow}
\usepackage[table,xcdraw]{xcolor}
\def\BibTeX{{\rm B\kern-.05em{\sc i\kern-.025em b}\kern-.08em
    T\kern-.1667em\lower.7ex\hbox{E}\kern-.125emX}}

\begin{document}

\title{Zero-Shot Classification of Crisis Tweets Using Instruction-Finetuned Large Language Models*\\
{
}
\thanks{This material is based upon work supported by the Department of the Air Force under Air Force Contract No. FA8702-15-D-0001. Any opinions, findings, conclusions or recommendations expressed in this material are those of the author(s) and do not necessarily reflect the views of the Department of the Air Force.
\textsuperscript{\textcopyright} 2024 Massachusetts Institute of Technology.
Delivered to the U.S. Government with Unlimited Rights, as defined in DFARS Part 252.227-7013 or 7014 (Feb 2014). Notwithstanding any copyright notice, U.S. Government rights in this work are defined by DFARS 252.227-7013 or DFARS 252.227-7014 as detailed above. Use of this work other than as specifically authorized by the U.S. Government may violate any copyrights that exist in this work.}
}

\author{\IEEEauthorblockN{Emma L. McDaniel\IEEEauthorrefmark{1}\IEEEauthorrefmark{2}, Samuel Scheele\IEEEauthorrefmark{1}\IEEEauthorrefmark{3}, Jeffrey Liu\IEEEauthorrefmark{4}}
\IEEEauthorblockA{\IEEEauthorrefmark{1} indicates equal contribution\\
\IEEEauthorrefmark{2}\textit{Computer Science Department} \\
\textit{Georgia State University}\\
Atlanta, GA, USA}
\IEEEauthorblockA{\IEEEauthorrefmark{2}\IEEEauthorrefmark{3}\IEEEauthorrefmark{4}\textit{Humanitarian Assistance and Disaster Relief Systems} \\
\textit{MIT Lincoln Laboratory}\\
Lexington, MA, USA \\
\texttt{\IEEEauthorrefmark{2}emcdaniel10@gsu.edu, \IEEEauthorrefmark{3}samuel.scheele@ll.mit.edu, \IEEEauthorrefmark{4}jeffrey.liu@ll.mit.edu}
}
}

\maketitle


\begin{abstract}
Social media posts are frequently identified as a valuable source of open-source intelligence for disaster response, and pre-LLM NLP techniques have been evaluated on datasets of crisis tweets. We assess three commercial large language models (OpenAI GPT-4o, Gemini 1.5-flash-001 and Anthropic Claude-3-5 Sonnet) capabilities in zero-shot classification of short social media posts. In one prompt, the models are asked to perform two classification tasks: 1) identify if the post is informative in a humanitarian context; and 2) rank and provide probabilities for the post in relation to 16 possible humanitarian classes. The posts being classified are from the consolidated crisis tweet dataset, CrisisBench. Results are evaluated using macro, weighted, and binary F1-scores. The informative classification task, generally performed better without extra information, while for the humanitarian label classification providing the event that occurred during which the tweet was mined, resulted in better performance. Further, we found that the models have significantly varying performance by dataset, which raises questions about dataset quality.
\end{abstract}

\begin{IEEEkeywords}
Large Language Models, Zero-Shot Classification, Crisis Classification, Social Media
\end{IEEEkeywords}

\section{Introduction}

In crisis scenarios, such as natural hazard-induced disasters or humanitarian emergencies, timely and accurate information is crucial to decision makers. Social media posts can provide valuable information in real time; however, the sheer speed and quantity of data coming from social media can be overwhelming for human analysts to process. As such, Natural Language Processing (NLP) techniques have been used to automate the processing of social media data in order to classify and extract the most relevant information. CrisisBench \cite{alam2020standardizing} provides a benchmark dataset to evaluate the performance of NLP solutions for classifying crisis-related social media posts. 

Recently, Large Language Models (LLMs) and Large Multimodal Models (LMMs) have shown impressive performance on a wide range of NLP tasks without needing task-specific training or fine-tuning. Large Language Models, such as GPT \cite{brown2020language, openai2024gpt4technicalreport}, are trained on massive text datasets to predict the next word in a sequence, and can be used to generate answers to questions. They can thus be used as zero-shot text classifiers by inputting the relevant text, followed by a question asking which of a set of given labels or classes apply to the text. Large Multimodal Models can be trained and utilized similarly, except they are configured to accept other data modalities, such as images, in addition to text. 

Due to the recent popularity of LLMs, we expect that humanitarian practitioners will try to use them to help automate the process of extracting relevant information from social media during crises. As a first step towards characterizing the performance of LLM/LMMs on such tasks and identifying which ones provide the best performance, we evaluate various open-access and commercial LLMs and LMMs on zero-shot classification of social media posts using the CrisisBench dataset. In addition, we compare the performance of the zero-shot classifiers to the existing benchmarks from purpose-built classifiers for crisis-related social media text classification.

\subsection{Related Work}

CrisisBench~\cite{alam2020standardizing} combines a number of crisis datasets \cite{imran2016lrec, crisismmd2018icwsm, olteanu2015expect, olteanu2014crisislex, imran2013practical, imran2013extracting} through cleaning and standardizing labels in order to create a benchmark for measuring performance of NLP classification of crisis-related social media posts. CrisisBench defines two tasks. The ``informativeness" task, a binary classification task that seeks to identify whether a provided tweet contains valuable information regarding a disaster or crisis event. The ``humanitarian information type," a multi-class classification task that seeks to categorize a tweet into one of 16 classes (e.g donation and volunteering, displaced and evacuations).

Previous work in classifying crisis-related social media posts has involved conventional machine learning methodologies and non-transformer-based neural networks~\cite{ofli2020analysis, li2021combining, mondal2022efficient, dasari2023stacking, asinthara2023categorizing,  krishna2024deep}, fine-tuned transformer-based models for multimodal classification using images embedded within social media messages~\cite{li2024mhrn}, and fine-tuned transformer-based models focusing solely on text~\cite{li2021combining, francois2023active, dahou2023social, malik2024categorization, Lamsal_2024}. 

In contrast to developing task-specific models, a growing trend involves leveraging instruction-tuned LLMs and LMMs for zero-shot classification~\cite{xian2018zero, Kojima2022Large, li2023revisiting, wu2023matching, Yu2023Zero, Zheng2023Judging, latif2024fine, sushil2024comparative}. While zero-shot classification circumvents the need for extensive labeled training data for fine-tuning, it is essential to understand various models' limitations within specific domains. Our work attempts to address this need in the realm of humanitarian assistance by providing performance statistics for a range of commercial models.

\section{Methodology}

\subsection{CrisisBench Task Descriptions}
In this paper, we focus on the ``informativeness" task from the CrisisBench consolidated dataset, and provide incidental analysis of the ``humanitarian information type" task for those data points that also had ``humanitarian information" labels. Roughly 5,000 of the examples in the informativeness test set are also in the humanitarian information type test set - these examples are the only ones considered for the analysis of the humanitarian information type task.

The motivation for this is that the classes in the multi-class task are often amalgamations of classes from the constituent datasets, and most constituent datasets used only a few of the classes. 
In the \textit{Discussion}, we will provide preliminary analysis of the multi-class task where semantic differences in definitions across constituent datasets substantially impacted performance. 
The ``informativeness" task does not suffer from the same ambiguity as the ``humanitarian information type" task and results are therefore easier to obtain and interpret.

A subset of tweets in the CrisisBench dataset are from CrisisMMD dataset, which contains only tweets which include images \cite{crisismmd2018icwsm}. The CrisisBench authors also include an event type annotation that indicates the type of crisis event that was contemporaneous with the timestamp of the tweet. We evaluate each task both with/without event awareness, and with/without images for four configurations per task. 

\subsection{Models Evaluated}
We evaluate three commercial models: OpenAI's GPT-4o \cite{gpt-4o}, Google's Gemini 1.5 Flash \cite{gemini-15-flash}, and Anthropic's Claude Sonnet 3.5 \cite{claude-35-sonnet}, and accessed them through their respective APIs. 
The models were chosen based on several considerations, including prominence, performance on other benchmarks, and availability.

\subsection{Prompt Structure}
We used the same base prompt for all models in the CrisisBench dataset, in which 
we requested that the model return a JSON string with a specified schema. We used Pydantic to validate the JSON. In the case where the model did not 
return valid JSON, we simply re-submitted the prompt and retried up to a set patience of three attempts. Responses that were not valid were omitted from analysis.

We asked the models to complete both the ``informativeness" task as well as the multi-class ``humanitarian information type" classification task in the same prompt. For the informativeness task the model provides a true or false
The base prompt is provided below,
\begin{lstlisting}
Provide classifications of the following tweet based on its relevance to a humanitarian event and a classification of its content. "\
        "{img_str}"\
        "{event_str}"\
        "The tweet follows:\n{tweet_str} \n"\
        "{field_descriptions}"
\end{lstlisting}
where the fields \texttt{\{img\_str\}}, \texttt{\{event\_str\}}, \texttt{\{tweet\_str\}}, and \texttt{\{field\_descriptions\}} are placeholders. 

The placeholder \texttt{\{img\_str\}} was filled in with the text \texttt{"Use the images, if present, to help you make a your determinations related to the informativeness and category of the tweet."} if an image was associated with the tweet; otherwise, it was left blank. Images were resized to fit within $768\times768$ pixels while maintaining aspect ratio, encoded in base64, and appended to the prompt in accordance to the respective LMM's specifications.

The placeholder \texttt{\{event\_str\}} was filled in with \texttt{"While it may still be irrelevant or uninformative, this tweet was created around the time of a disaster with description: \{event\_type\}."}, where \texttt{\{event\_type\}} corresponds to the event that was occurring during the time of the tweet if we were evaluating the tweet in the ``event-aware" configuration; otherwise, \texttt{\{event\_str\}} was left blank. The \texttt{\{tweet\_str\}} placeholder contained the actual text of the tweet.

The \texttt{\{field\_descriptions\}} placeholder contained descriptions of the classes as well as the desired JSON format for the output.
We requested two fields: \texttt{is\_informative} and \texttt{humanitarian\_label}. For the \texttt{is\_informative} field, the prompt was \texttt{"Does the tweet contain information pertinent to a humanitarian event or natural disaster? Respond with a boolean true/false"}. For the \texttt{humanitarian\_label} field, the prompt is provided below:
\begin{lstlisting}
For a given tweet, determine which of the humanitarian labels are most relevant:
The humanitarian labels and their descriptions are provided below:
    "not_humanitarian" - The tweet is not humanitarian in nature and does not fit into any other class.
    "donation_and_volunteering" - The tweet relates to directing, accepting, or distributing donations or volunteer effort.
    "requests_or_needs" - The tweet describes a request or need of an individual or community.
    "sympathy_and_support" - The tweet expresses sympathy or support for disaster victims.
    "infrastructure_and_utilities_damage" - The tweet relates to the construction or destruction of infrastructure, utilities, or structures. 
    "affected_individual" - The tweet contains information on a particular individual affected by a disaster.
    "caution_and_advice" - The tweet contains caution or advice for victims, responders, or others.
    "injured_or_dead_people" - The tweet notes the presence of injured or dead people.
    "response_efforts" - The tweet pertains to the response effort.
    "missing_and_found_people" - The tweet discusses missing persons, including in the context of finding them.
    "displaced_and_evacuations" - The tweet relates to displaced people or an evacuation process.
    "personal_updates" - The tweet relates to a personal opinion or a status update about the tweet author or their close relations.
    "physical_landslide" - The tweet is related to a physical landslide.
    "disease_related" - The tweet reports on disease transmissions, symptoms, treatment, prevention, or affected people.
    "terrorism_related" - The tweet reports possible terrorism or terrorist acts.
    "other_relevant_information" - The tweet is humanitarian in nature, but does not fit in any other class.
The output should be formatted as a dictionary whose keys are the humanitarian labels, and the values are two-element arrays whose entries correspond to the following:
The first element of the array should be a ranking: an integer from 1 to 16 representing the relative relevance of the humanitarian label compared to the others. The most relevant label should be ranked 1, and least relevant should be ranked 16.
The second element should be a likelihood score: a floating point number between 0 and 1 representing the likelihood that the label applies to the tweet.
The dictionary should have an entry for every humanitarian label, even if it is not relevant. The rankings for each label should be unique---that is, no two labels should have the same ranking, even if they are both not relevant: you must rank one higher than the other---and the likelihoods should sum to 1
\end{lstlisting}

At the end of the prompt, we provided an example of the JSON format to address errors encountered during models' JSON construction and subsequent validation using Langchain/Pydantic:
\begin{lstlisting}
For example, a correctly formatted answer would be: {"is_informative": false, "humanitarian_label": {"not_humanitarian": [1, 0.95], "donation_and_volunteering": [16, 0.005], "requests_or_needs": [15, 0.005], "sympathy_and_support": [2, 0.01], "infrastructure_and_utilities_damage": [14, 0.002], "affected_individual": [13, 0.002], "caution_and_advice": [12, 0.002], "injured_or_dead_people": [11, 0.002], "response_efforts": [10, 0.002], "missing_and_found_people": [9, 0.002], "displaced_and_evacuations": [8, 0.002], "personal_updates": [3, 0.008], "physical_landslide": [7, 0.002], "disease_related": [6, 0.002], "terrorism_related": [5, 0.002], "other_relevant_information": [4, 0.007]}
\end{lstlisting}

\begin{table}[b]
\centering
\caption{Summary of F1 Scores for Informativeness Task}
\begin{tabular}{lccc}
\toprule
&\multicolumn{1}{r}{\textit{Event Aware}} & \multicolumn{1}{c|}{} & \multicolumn{1}{c}{x} \\
\cline{3-4}
\cline{2-4}
\textit{Model} & \textit{Metric} &  &  \\
\multicolumn{1}{l}{Claude-3-5 Sonnet} & \texttt{macro} &  0.800 & 0.796 \\
& \texttt{binary} & 0.836 & 0.836  \\
& \texttt{weighted} & 0.808 & 0.805  \\
\cline{2-4}

\multicolumn{1}{l}{Gemini 1.5-flash-001}& \texttt{macro} & 0.802 & 0.799  \\
& \texttt{binary} & 0.855 & \textbf{0.848} \\
& \texttt{weighted} & 0.814 & \textbf{0.810 } \\
\cline{2-4}

\multicolumn{1}{l}{GPT-4o}& \texttt{macro} & \textbf{0.819} & \textbf{0.801} \\
& \texttt{binary} & \textbf{0.860} & 0.837 \\
& \texttt{weighted} & \textbf{0.828} & 0.809  \\
\bottomrule
\end{tabular}
\label{tab:inf_all}
\end{table}

\begin{table*}[!htbp]
\caption{F1 Scores for Informativeness Task by Each Dataset Across Commercial Models with and without Event Awareness}
\resizebox{\textwidth}{!}{
\begin{tabular}{rccccccccccccccccccc}
\toprule
&  & \multicolumn{2}{c}{CrisisLex6} & \multicolumn{2}{c}{CrisisLex26} & \multicolumn{2}{c}{CrisisNLP-cf} & \multicolumn{2}{c}{CrisisNLP-vol} & \multicolumn{2}{c}{AIDR} & \multicolumn{2}{c}{DSM} & \multicolumn{2}{c}{DRD} & \multicolumn{2}{c}{ISCRAM2013} & \multicolumn{2}{c}{SWDM13} \\ 
\cline{3-20} 
&\textit{Event Aware}& \multicolumn{1}{c|}{} & \multicolumn{1}{c|}{x} & \multicolumn{1}{c|}{} & \multicolumn{1}{c|}{x} & \multicolumn{1}{c|}{} & \multicolumn{1}{c|}{x} & \multicolumn{1}{c|}{} & \multicolumn{1}{c|}{x} & \multicolumn{1}{c|}{} & \multicolumn{1}{c|}{x} & \multicolumn{1}{c|}{} & \multicolumn{1}{c|}{x} & \multicolumn{1}{c|}{} & \multicolumn{1}{c|}{x} & \multicolumn{1}{c|}{}  & \multicolumn{1}{c|}{x} & \multicolumn{1}{c|}{} & \multicolumn{1}{c}{x} \\
\cline{3-20} 

\multicolumn{1}{l}{\textit{Model}}&\textit{Metric}&  &  &  &  &  &  &  &  &  &  &  &  &  &  &  &  &  &  \\
\multicolumn{1}{l}{Claude-3-5 Sonnet} & \texttt{macro} & 0.840 & 0.825 & 0.603 & \textbf{0.617} & 0.759 & 0.739 & 0.667 & 0.687 & \textbf{0.746} & 0.739 & 0.728 & 0.716 & 0.740 & \textbf{0.743} & 0.498 & 0.481 & 0.555 & 0.557 \\

& \texttt{binary} & 0.812 &  0.796 &  0.916 &  0.932 &  0.936 &  0.921 &  0.640 &  0.655 &  \textbf{0.707} &  0.693 &  0.617 &  0.599 &  0.839 &  0.841 &  0.892 &  0.871 &  0.765 &  0.777 \\
&\texttt{weighted} & 0.842 & 0.827 & 0.886 & 0.902 & 0.896 & 0.881 & 0.676 & 0.697 & \textbf{0.752} & 0.746 & 0.753 & 0.743 & 0.782 & 0.785 & 0.869 & 0.849 & 0.689 & 0.697 \\
\cline{2-20} 

\multicolumn{1}{l}{Gemini 1.5-flash-001} & \texttt{macro} & 0.860 & \textbf{0.834} & 0.590 & 0.600 & 0.744 & 0.738 & 0.656 & 0.689 & 0.718 & 0.725 & \textbf{0.787} & \textbf{0.774} & 0.687 & 0.695 & \textbf{0.560} & \textbf{0.536} & 0.588 & \textbf{0.598} \\

& \texttt{binary} & 0.842 &  \textbf{0.814} &  \textbf{0.943} &  \textbf{0.940} &  0.941 &  \textbf{0.930} &  0.639 &  0.658 &  0.706 &  0.699 &  \textbf{0.731} &  \textbf{0.708} &  0.849 &  \textbf{0.851} &  \textbf{0.951} &  \textbf{0.927} & \textbf{ 0.848} &  \textbf{0.846} \\
&\texttt{weighted} & 0.861 & \textbf{0.835} & 0.909 & \textbf{0.908} & 0.897 & \textbf{0.887} & 0.662 & 0.699 & 0.719 & 0.729 &  \textbf{0.800} & \textbf{0.789 }& 0.756 & 0.761 & \textbf{0.929} & \textbf{0.905} & \textbf{ 0.753} & \textbf{0.756} \\
\cline{2-20} 

\multicolumn{1}{l}{GPT-4o}  & \texttt{macro}  & \textbf{0.879} & 0.822 & \textbf{0.622} & 0.610 & \textbf{0.786} &\textbf{ 0.748} & \textbf{0.670} & \textbf{0.726} & 0.738 & \textbf{0.741} & 0.744 & 0.732 & \textbf{0.753} & 0.735 & 0.524 & 0.456 & \textbf{0.591} & 0.543 \\

&\texttt{binary} & \textbf{0.862} &  0.789 &  0.940 &  0.922 & \textbf{0.950} &  0.923 & \textbf{0.645} &  \textbf{0.682} &  0.702 &  \textbf{0.701} &  0.644 &  0.624 &  \textbf{0.855} &  0.847 &  0.926 &  0.836 &  0.822 &  0.743 \\
&\texttt{weighted} & \textbf{0.880} & 0.824 & \textbf{0.910} & 0.892 & \textbf{0.913} & 0.884 & \textbf{0.678} & \textbf{0.740} & 0.743 & \textbf{0.747} & 0.766 & 0.756 & \textbf{0.796} & \textbf{0.782} & 0.903 & 0.815 & 0.738 & 0.670 \\

\bottomrule
\end{tabular}}
\label{tab:inf_allmone_datasets}
\end{table*}

\begin{table}[!htbp]
\caption{F1 Scores for Informativeness Task on CrisisMMD with/without Event Awareness and Use of Images}
\centering
\begin{tabular}{rccccc}
\toprule
&\textit{Event Aware}& \multicolumn{2}{c|}{} & \multicolumn{2}{c}{x} \\ 
\cline{3-6} 
&\textit{Image Used}& \multicolumn{1}{c|}{} & \multicolumn{1}{c|}{x} & \multicolumn{1}{c|}{} & \multicolumn{1}{c}{x} \\ 
\cline{3-6} 
\multicolumn{1}{l}{\textit{Model}} & \textit{Metric} &  &  &  &  \\
\multicolumn{1}{l}{Claude-3-5 Sonnet} & \texttt{macro} & \textbf{0.760} & \textbf{0.712} & 0.743 & 0.\textbf{731} \\
 & \texttt{binary} & \textbf{0.877} &  0.869 & \textbf{ 0.871} &  \textbf{0.871}\\
 & \texttt{weighted} & \textbf{0.809} & \textbf{0.776} & 0.795 & \textbf{0.788} \\
\cline{2-6} 
\multicolumn{1}{l}{Gemini 1.5-flash-001} & \texttt{macro} & 0.701 & 0.703 & 0.745 & 0.729 \\
 & \texttt{binary} & 0.873 & \textbf{ 0.872} &  0.870 &  0.869\\
 & \texttt{weighted} & 0.771 & 0.772 & 0.796 & 0.786 \\
\cline{2-6} 
\multicolumn{1}{l}{GPT-4o} & \texttt{macro} & 0.733 & 0.699 & \textbf{0.761} & 0.728 \\
& \texttt{binary} & 0.876 &  0.870 &  0.868 &  0.870\\
 & \texttt{weighted} &0.791 & 0.769 & \textbf{0.805} & 0.786 \\
\bottomrule
\end{tabular}
\label{tab:inf_crisismmd_datasets}
\end{table}

\begin{table*}[!htbp]
\caption{Macro and Weighted F1 Scores for Humanitarian Classification by each Dataset}
\resizebox{\textwidth}{!}{
\begin{tabular}{llrrrrrrrrrrrrrrrrrrrr}
\toprule
& \multicolumn{1}{r}{\textit{Dataset}} & \multicolumn{4}{c}{CrisisMMD} & \multicolumn{2}{c}{CrisisLex6} & \multicolumn{2}{c}{CrisisLex26} & \multicolumn{2}{c}{CrisisNLP-cf} & \multicolumn{2}{c}{CrisisNLP-vol} & \multicolumn{2}{c}{AIDR} & \multicolumn{2}{c}{DRD} & \multicolumn{2}{c}{ISCRAM2013} & \multicolumn{2}{c}{SWDM13} \\
\cline{3-22} 
& \multicolumn{1}{r}{\textit{Event Aware}} & \multicolumn{2}{c|}{} & \multicolumn{2}{c|}{x} &\multicolumn{1}{c|}{}&\multicolumn{1}{c|}{x}&\multicolumn{1}{c|}{}&\multicolumn{1}{c|}{x}&\multicolumn{1}{c|}{}&\multicolumn{1}{c|}{x}&\multicolumn{1}{c|}{}&\multicolumn{1}{c|}{x}&\multicolumn{1}{c|}{}&\multicolumn{1}{c|}{x}&\multicolumn{1}{c|}{}&\multicolumn{1}{c|}{x}&\multicolumn{1}{c|}{}&\multicolumn{1}{c|}{x}&\multicolumn{1}{c|}{}&\multicolumn{1}{c}{x}\\
\cline{3-22} 
&\multicolumn{1}{r}{\textit{Image Used}} &\multicolumn{1}{c|}{}&\multicolumn{1}{c|}{x}&\multicolumn{1}{c|}{}&\multicolumn{1}{c}{x}&\multicolumn{1}{c}{}&  &\multicolumn{1}{c}{}&  &\multicolumn{1}{c}{}&  &\multicolumn{1}{c}{}&  &\multicolumn{1}{c}{}&  &\multicolumn{1}{c}{}&  &\multicolumn{1}{c}{}&  &\multicolumn{1}{c}{}&  \\
\cline{3-6} 
\textit{Model} & \textit{Metric} &&  &&  &&  &&  &&  &&  &&  &&  &&  &&  \\
Claude-3-5 Sonnet & \texttt{macro} & \textbf{0.537} & \textbf{0.508} & \textbf{0.554} & 0.493 & 0.836 & 0.832 & \textbf{0.461} & \textbf{0.492} & 0.564 & 0.553 & \textbf{0.263} & 0.250 & 0.345 & 0.333 & \textbf{0.700} & \textbf{0.680} & 0.530 & \textbf{0.547} & 0.486 & \textbf{0.540} \\
& \texttt{weighted} & \textbf{0.625} & \textbf{0.557} & \textbf{0.638} & 0.558 & 0.837 & 0.834 & \textbf{0.467} & \textbf{0.497} & 0.583 & 0.570 & 0.546 & 0.558 & 0.623 & 0.612 & \textbf{0.837} & \textbf{0.823} & \textbf{0.653} & \textbf{0.618} & 0.590 & \textbf{0.666} \\
\cline{2-22} 
Gemini 1.5-flash-001 & \texttt{macro} & 0.500 & 0.466 & 0.533 & 0.472 & 0.869 & \textbf{0.838} & 0.439 & 0.445 & 0.546 & 0.540 & 0.193 & 0.219 & \textbf{0.363} & \textbf{0.346} & 0.488 & 0.501 & \textbf{0.541} & 0.526 & 0.440 & 0.440 \\
& \texttt{weighted} & 0.531 & 0.510 & 0.565 & 0.527 & 0.869 & \textbf{0.839} & 0.438 & 0.452 & 0.565 & 0.549 & 0.540 & 0.598 & \textbf{0.637} & \textbf{0.626} & 0.617 & 0.633 & 0.610 & 0.581 & 0.549 & 0.549 \\
\cline{2-22} 
GPT-4o & \texttt{macro} & 0.513 & 0.485 & 0.531 &\textbf{ 0.504 }& \textbf{0.882} & 0.822 & 0.448 & 0.464 & \textbf{0.603} & \textbf{0.583 }& 0.254 & \textbf{0.262} & 0.341 & 0.300 & 0.627 & 0.611 & 0.532 & 0.538 & \textbf{0.505} & 0.433 \\
& \texttt{weighted} & 0.568 & 0.545 & 0.586 & \textbf{0.563} & \textbf{0.882} & 0.824 & 0.451 & 0.468 & \textbf{0.626} & \textbf{0.585} & \textbf{0.596} & \textbf{0.644} & 0.625 & 0.605 & 0.773 & 0.754 & 0.645 & 0.612 &\textbf{ 0.624} & 0.537 \\

\bottomrule
\end{tabular}}
\label{tab:humtable}
\end{table*}

\subsection{Evaluation}
For both the informativeness and humanitarian label tasks, 
we calculate F1 scores to facilitate comparison with existing evaluations of datasets within CrisisBench. For the informativeness classification, we calculate macro (unweighted), weighted, and binary (only the positive class) F1 scores; for the humanitarian classification, we calculate weighted and macro (unweighted) class-averaged F1 scores. Given that some datasets in CrisisBench may cover only a subset of the 16 labels specified in the prompt, we maintain consistency in the prompt by including all 16 labels in our evaluation framework. However, the performance assessment of each dataset is based exclusively on the rankings of the labels present within that dataset.
Class-specific precision for class $i$ is computed as $\text{P}_i = \frac{\text{TP}_i}{\text{TP}_i + \text{FP}_i}$, where $\text{TP}_i, \text{FP}_i$ stand for the count of true positives and false positives for class $i$, respectively. Class-specific recall is defined as $\text{R}_i = \frac{\text{TP}_i}{\text{TP}_i + \text{FN}_i}$, where $\text{FN}_i$ is the count of false negatives for class $i$. Class-specific F1 is defined as 
\begin{equation*}
    F1_i = \frac{2\text{P}_i\text{R}_i}{\text{P}_i+\text{R}_i}
\end{equation*}
In the case where there is only one class, the class-specific F1 is equivalent to the binary F1. 

The weighted class-average F1 is defined as
\begin{equation*}
    F1_\text{weighted} = \sum_i \frac{n_i}{N} \text{F1}_i
\end{equation*}
where $n_i$ is the number of instances where the true class is $i$, and $N$ is the total number of instances. The macro class-average F1 is defined as
\begin{equation*}
F1_\text{macro} = \sum_i \frac{1}{m} \text{F1}_i
\end{equation*} where $m$ is the number of classes.

\section{Results}

In our experiment, the classification tasks for the crisis tweets on informativeness and type of humanitarian label are combined into one prompt. The model was prompted to: 1) determine if the social media post is informative in a humanitarian context, and 2) rank and assign probabilities to 16 potential humanitarian labels in how it fits the post. We report macro, weighted, and binary F1 scores for the informativeness task. For the humanitarian task, we use macro and weighted F1 scores.

Results for the informativeness task across the three models for all tested crisis tweets are in Table~\ref{tab:inf_all}. Weighted, binary and macro F1 scores are reported with and without event awareness. 
``Event-aware" indicates whether the name of a disaster contemporaneous with the tweet is included in the prompt. For each version of the prompt, the highest F1 scores are bolded. Models performed slightly better without event-awareness. OpenAI's GPT-4o performed the best without event-awareness at a macro F1 score of 0.819, a binary F1 score at 0.860, and weighted F1 score at 0.828.

The results by dataset are in two tables, Table~\ref{tab:inf_allmone_datasets} contains all datasets not including CrisisMMD, and Table~\ref{tab:inf_crisismmd_datasets} contains the CrisisMMD results. These results are also separated by whether extra information (event and/or image) was included in the prompt. Across most datasets, OpenAI GPT-4o outperformed the other evaluated LLMs in informativeness classifications. 

Based on reported metrics, the best LLMs compares moderately lower to existing benchmarks on the consolidated dataset \cite{alam2020standardizing}, with a weighted F1 of 0.828 (LLM: GPT-4o) vs. 0.883 (fine-tuned RoBERTa). When looking at specific datasets where benchmarks were available, the LLMs also underperform, sometimes by a large margin: on CrisisMMD \cite{ofli2020analysis}, (weighted F1 of .638 vs .842), on CrisisLexT26 \cite{li2021combining} (macro F1 of .492 vs .848), and CrisisLexT6 \cite{li2021combining} (macro F1 of .882 vs .947). 

As a note, we contend that binary F1 (class-wise F1 for single class) as the most appropriate metric for this task. Macro and weighted F1 are most appropriate metrics for multi-class classification, in which F1 scores for multiple classes are condensed into a single metric. For this binary classification problem, the ``informative" class serves as a foreground and the ``not informative" class as a background; binary F1 appropriately privileges the foreground class as being the one which is useful to distinguish. The use of macro and weighted F1 are significantly impacted by the number of negative examples in the dataset, which is not desirable when the task is trying to find positive instances in a haystack of negative examples. The LLMs obtain substantially better binary F1 scores than macro F1. Unfortunately, binary F1 is not reported for the informativeness task in any literature we found, making direct comparison on the metric difficult. 

For the informativeness task, the inclusion of images (in CrisisMMD) and event context (all datasets) did not significantly affect F1 scores, with changes dependent on the dataset. For humanitarian classification, including event context slightly improved scores, while including images did not.

For the humanitarian classification task, we compute both weighted and macro F1 scores, and treat the highest-ranked label within the subset of labels from the constituent dataset as the predicted label for evaluation. These results are reported in Table~\ref{tab:humtable}. The highest F1 score for both macro and weighted for each dataset are bolded for each variation of the prompt. Anthropic Claude-3-5 Sonnet generally performed better than other tested LLMs in both weighted and macro f1-scores. The F1 scores of the LLMs are broadly low, but inconsistencies in the CrisisBench dataset are likely substantially responsible for this fact, which we will explore further in the discussion. We also note that this evaluation of humanitarian labels is preliminary, as it only includes the subset of tweets in the informativeness task that also happened to have humanitarian label annotations.

\begin{table}[b]
\caption{Accuracy rates of two classes for Anthropic's Claude-3.5 Sonnet and manually assigned labels compared to ground-truth}
\resizebox{.5\textwidth}{!}{
\begin{tabular}{l|c|c}
Label & Accuracy (ground-truth) \% & Accuracy (manual) \% \\
\midrule
\texttt{affected\_individual} & 13.3 & 76.0 \\
\texttt{caution\_and\_advice} & 14.7 & 48.0 \\
\texttt{disease\_related} & 56.0 & 74.7 \\
\texttt{affected\_individual} (ground-truth) & 100 & 29.3 \\
\texttt{caution\_and\_advice} (ground-truth) & 100 & 61.3 \\
\texttt{disease\_related} (ground-truth) & 100 & 65.3
\end{tabular}
}

\label{tab:manual-analysis}
\end{table}

\section{Discussion}
Overall, LLMs perform reasonably well on the informativeness task, achieving zero-shot performance on the consolidated dataset within 6\% of that of pretrained classifiers in \cite{alam2020standardizing}. 

However, performance was substantially worse on the multi-class humanitarian label task. The LLMs broadly underperformed the models trained in CrisisTransformers \cite{Lamsal_2024}. We note several limitations and challenges that may have contributed to this. The task performance of the LLMs tested may have been limited by a number of factors, including the absence of optimizations like prompt engineering or fine-tuning. Further, fine-tuned models may perform better because they were fine-tuned on each dataset individually and were able to fit the base rates at which various classes occur. More fundamentally, however, we note that methods in the construction of CrisisBench dataset had a substantial impact on the multi-class task performance for the LLMs.

Whereas the zero-shot classification technique for LLMs uses natural language prompting to describe the criteria for each class based on the name and description of the label, traditional models are trained on the training dataset. This raises a potential issue if there is misalignment between the labeled examples and the semantic understanding of the label. We identified methods related to the construction of CrisisBench which may contribute to such misalignment.

CrisisBench draws from a number of datasets, aggregating labels with potentially different definitions into single classes. For example, the \texttt{infrastructure\_and\_utilities\_damage} class is defined in CrisisNLP-volunteers to be the destruction of houses, buildings, or roads, or the interruption of utilities, but CrisisNLP-CF defines it to include \textit{restoration} of utilities as well \cite{alam2020standardizing}. 
Furthermore, the classes of constituent datasets are sometimes defined such that they cannot be mapped onto a single CrisisBench label. The CrisisLexT26 \texttt{affected\_individual} class contributes all of the \texttt{affected\_individual} examples in CrisisBench. But in CrisisLexT26, this class is defined to include personal updates, which is a separate CrisisBench class. The only way for an LLM to correctly categorize a personal update is to correctly guess whether it came from CrisisLexT26 or not. As CrisisLexT26 is one of the largest datasets labeled for humanitarian class, it is unsurprising that \texttt{personal\_update} and \texttt{affected\_individual} are the two lowest-performing classes for the evaluated LLMs.

To better understand the impact of ambiguous label mappings on performance, we performed manual binary annotation on two of the lowest-performing classes, \texttt{affected\_individual} and \texttt{caution\_and\_advice}. We examined 75 randomly sampled tweets which were assigned to the two classes (for a total of 150 tweets) by either the ground truth label or the maximum likelihood estimate of the LLM, without knowledge of the ground-truth or predicted label of any particular tweet. We manually performed binary classification on each tweet as either matching or not matching the description of its reference class given in the prompt (for example, we might look at a tweet with the knowledge that at least one of the ground truth label or the predicted label was \texttt{affected\_individual}, and assign a binary label based on whether we believed the tweet matched the definition we gave for that class). The results of this experiment are in Table~\ref{tab:manual-analysis}. This experiment suggests that semantic differences in the labels, which would not have affected models trained on the training data~\cite{Lamsal_2024}, had a substantial impact on the performance of the LLMs. It also showed generally low agreement between our manual labels and the ground truth, with our manual labels matching the LLM labels more often than the ground truth on the \texttt{affected\_individuals} class. While the LLMs performed better on our manual labels than on the ground truth in general, the difference in performance is much larger in the cases where agreement between our labels and the ground truth was relatively weaker. To verify that we were not observing a regression to the mean by injecting noise into the labels, we also analyzed a further 75 tweets from a high-performing class, \texttt{disease\_related}. Even on this higher-performing class, accuracy is substantially better when comparing against manually labeled examples as opposed to the ground truth labels. We also observe that accuracy as measured against the ground truth is higher on labels where the ground truth and manual labeling have higher agreement.


The significant level of variation in F1 scores between datasets across both classification tasks merits further investigation. For example, the binary F1 Score for OpenAI's GPT-4o of 0.950 suggests strong performance for the positive label on the CrisisNLP-cf dataset (labeled by paid crowd workers), and much worse performance with 0.682 on the CrisisNLP-vol dataset (labeled by unpaid volunteers). One possible explanation for this is that data quality varies substantially between constituent datasets.

We also note a couple additional challenges during implementation that practitioners and researchers using LLMs should be aware of. Occasionally, LLMs would refuse to classify a tweet due to objectionable subject material (pornographic content, hate speech, and foul language). In addition, LLMs sometimes struggled to output correctly formatted JSON---this was usually able to be resolved by resubmitting the prompt, but on rare occasion, the request failed past our patience threshold and had to be omitted. There were also a small number of tweets that were misconstrued as part of the prompt, leading the LLM to respond that it did not detect a tweet to classify. A stronger distinction between the prompt and tweet to be classified, perhaps using a special token, would help ameliorate this. The total number of refused/omitted tweets were small (on the order of 10 per model), and thus should not affect the evaluation scores. 

\section{Conclusion}

In this paper, we present the performance of commercial large language models on zero-shot classification for two tasks on short social media posts on CrisisBench. We find that overall performance on the binary informativeness task is strong, even relative to models fine-tuned on the evaluation datasets. Incorporating extra information in the form of possible event context and images did not substantially impact the model's performance on the task. 

For the second task, humanitarian classification, an ambiguous multi-class task performance rapidly declined, emphasizing the need for careful deployment of these tools to the humanitarian space. Based on small-scale experiments with manual labeling, we attribute most of the LLMs' declining performance to semantic ambiguity in social media posts and their labels rather than a latent inability to parse and classify natural language. 

In future work, we plan to include open-source models in our classification assessments. We will substantially reduce problems associated with the dataset aggregation performed by CrisisBench by changing the prompt and label definitions based on source dataset. This will also provide an avenue for further analysis of each dataset's quality. Further, we plan to analyze the results by language to better understand the multi-lingual components of the LLMs in relation to humanitarian classification tasks. Another avenue of future research is to assess the impact of prompt engineering more broadly. For example, we prompt for both classification tasks in the same prompt, but it would be of interest to look into the extent to which the dual classification task in one prompt impacts model performance.


\Urlmuskip=0mu plus 1mu
\interlinepenalty=10000 
\printbibliography
\end{document}